\documentclass{article}

\usepackage{microtype}
\usepackage{graphicx}
\usepackage{subcaption}
\usepackage{booktabs} 
\usepackage{booktabs}
\usepackage[most]{tcolorbox}

\usepackage{hyperref}

\usepackage[preprint]{icml2026}

\usepackage{amsmath}
\usepackage{amssymb}
\usepackage{mathtools}
\usepackage{amsthm}
\usepackage{multirow}

\usepackage[capitalize,noabbrev]{cleveref}

\theoremstyle{plain}

\theoremstyle{definition}

\theoremstyle{remark}

\usepackage[textsize=tiny]{todonotes}

\icmltitlerunning{Clarification is Not Enough: Post-Clarification Answering Remains the
Bottleneck in Multi-Turn QA}

\begin{document}

\twocolumn[
  \icmltitle{Clarification Is Not Enough: Post-Clarification Answering Remains the Bottleneck in Multi-Turn QA}




  \begin{icmlauthorlist}
    \icmlauthor{Jinyan Su}{a}
    \icmlauthor{Jennifer Healey}{b}

  \end{icmlauthorlist}

  \icmlaffiliation{a}{Cornell University, }
  \icmlaffiliation{b}{Adobe Research}
  \icmlcorrespondingauthor{Jennifer Healey}{jehealey@adobe.com}

  \icmlkeywords{Machine Learning, ICML}

  \vskip 0.3in
]



\printAffiliationsAndNotice{Work done when Jinyan was an intern at Adobe.\\}  

\begin{abstract}
Pluralistic alignment requires systems to adapt to diverse user values, communication styles, and contextual assumptions. We believe that a foundational prerequisite for such alignment enabling accurate preference elicitation from people when their intent is under-specified or ambiguous. We study the problem of preference elicitation in multi-turn question answering by decomposing the problem into two components: a \textbf{clarification policy}, which decides whether to ask a clarifying question or answer directly, and \textbf{post-clarification answering}, which produces the correct final answer once the missing information is provided. We show, using the PACIFIC benchmark, that supervised fine-tuning rapidly improves the clarification policy,  however, final answer accuracy remains substantially lower even when the model takes the correct action. This gap indicates that understanding and correctly interpreting the user's response is the critical gap in multi-turn question-answering systems. 
\end{abstract}

\section{Introduction}
While large language models (LLMs) have enabled conversational assistants that are increasingly fluent and capable across a wide range of tasks, a fundamental challenge remains: everyday language is often under specified and conversations often require clarification \cite{zipf2016human,piantadosi2012communicative, herlihy2024overcoming}. 
Users routinely omit key details, rely on assumed contextual understandings, and ask questions whose intended meaning depends on latent factors such as time, entity, scope, or goal. Such under-specification is can lead to efficient human communication, especially between people with a shared perspective, however, it creates a tension that AI assistants must navigate. The assistant must decide to either answer without clarifying which can lead to either overly verbose generic answers or incorrect answers or to ask one or more clarifying questions which might annoy the user.
\begin{figure*}[h]
    \centering \includegraphics[width=0.7\linewidth]{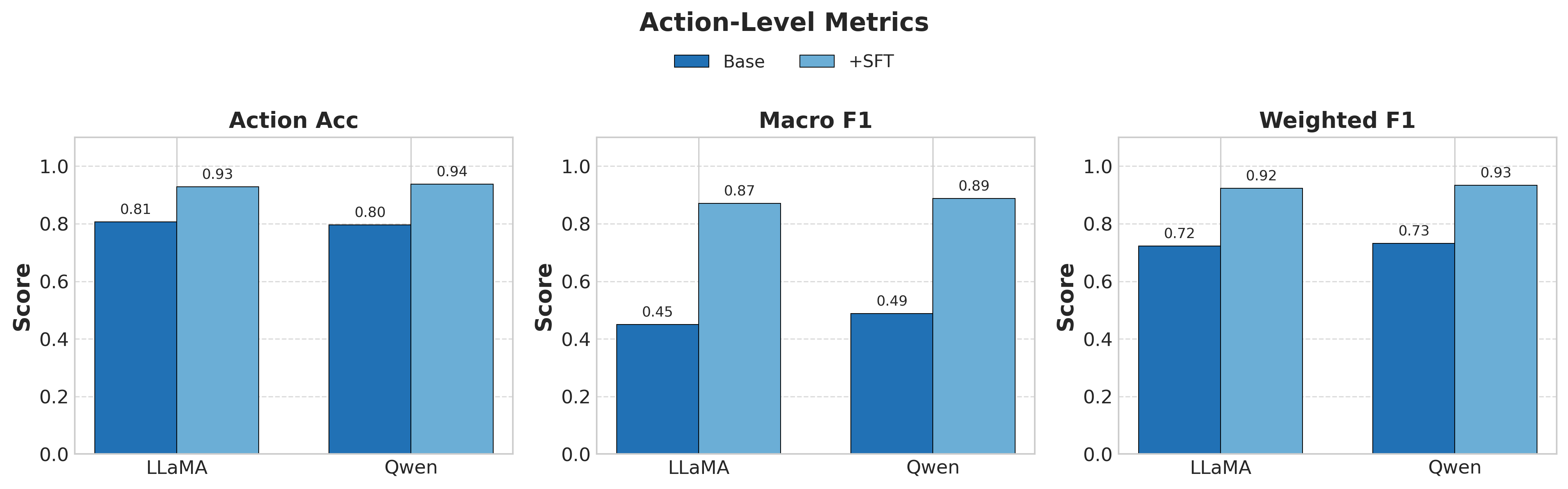}
    \caption{Action-level metrics (Action Acc, Macro F1, and Weighted F1) for LLaMA and Qwen before and after SFT. SFT substantially improves clarification policy for both models, with the largest gains in Macro F1.}
    \label{fig:action-level-metrics}
    \vspace{-0.5cm}
\end{figure*}
For example, when a user asks, “I’m planning a trip, what should I pack?”, the appropriate answer depends on missing information such as the destination, season, and planned activities. A standard assistant may respond immediately with an over-generalized checklist misaligned with the user's actual needs, or unnecessarily verbose. In contrast, an assistant that first asks a clarifying question can elicit the missing context and then provide a concise, situation-specific answer. Handling under-specification therefore requires not only fluent generation, but also the ability to manage uncertainty through interaction.

In this work, we decompose this ability into two components. The first is a \textbf{clarification policy}: deciding whether the next action should be to ask a clarifying question or to answer directly. The second is \textbf{post-clarification answering}: producing the correct final answer after the missing information has been provided. Prior work has largely focused on improving clarification behavior, but it remains unclear which of these two components is the main bottleneck in end-to-end multi-turn performance.

To study this question, we conduct controlled fine-tuning experiments on PACIFIC \citep{deng2022pacific}, a benchmark of naturally occurring multi-turn financial question answering containing both ambiguous and unambiguous user query turns. 
We fine-tune two instruction-tuned models: Meta-Llama-
3-8B-Instruct and Qwen2.5-7B-Instruct, and evaluate them with (i) action-level metrics, which measure whether the model chooses the correct clarification policy action (ask vs.\ answer), and (ii) final-answer metrics, which measure whether the model ultimately produces the correct answer. Crucially, we also report final-answer accuracy \emph{conditioned on taking the correct action}, which isolates failures in post-clarification answering from failures in clarification policy.

Our findings show a clear separation between these two components. Supervised fine-tuning rapidly improves the clarification policy, yielding strong gains in ask vs. answer decisions on both ambiguous and unambiguous turns, however, final answer accuracy remains substantially lower even when the model takes the correct action. This gap indicates that the main remaining bottleneck is not deciding \emph{whether} to clarify, but answering correctly \emph{after} clarification combining the response in the user's disambiguating information and the dialogue context. Together, these results suggest that progress on under-specified multi turn QA will require more than better clarification policies alone: improving post-clarification answering is equally, if not more, important.

\section{Related Work}
\subsection{Ambiguity and clarifying questions in QA}
Real-world user queries are often ambiguous or under-specified, which has motivated systems that ask clarifying questions before answering \citep{braslavski2017you, radlinski2017theoretical}. In information retrieval, prior work has shown that a well-chosen clarification question can substantially improve downstream retrieval quality. For example, \citet{aliannejadi2019asking} introduced the Qulac benchmark, a large-scale resource for studying clarifying questions over ambiguous queries, and showed that clarification can improve retrieval effectiveness. Subsequent work extended this line of research to conversational settings, where agents may ask one or more follow-up questions to refine user intent \citep{aliannejadi2020convai3}.

To support training and evaluation, several datasets have been developed from real or crowd-sourced queries, including MIMICS \citep{zamani2020mimics}, ClarQ \citep{kumar2020clarq}, CLAQUA \citep{xu2019asking}, and Abg-CoQA \citep{guo2021abg}. Across these benchmarks, clarification is typically framed as identifying ambiguity, asking for missing information, and then answering once the ambiguity is resolved. Our work builds on this line of research, but focuses specifically on multi-turn settings where ambiguity may arise not only from a single user query, but also from conversational history and follow-up context.

\subsection{Training LLMs to ask clarifying questions}
A large body of work has focused on teaching systems when to ask clarifying questions and what questions to ask. Prior studies have examined the decision of whether clarification is needed \citep{hancock2019learning, kim2021deciding}, as well as the generation of informative clarification questions \citep{rao2018learning, rao2019answer, wang2018learning}. More recent work studies clarification in the context of large language models, which often default to answering one plausible interpretation of an ambiguous query instead of explicitly seeking missing information \citep{zhangmodeling}. To address this, recent methods use future dialogue outcomes, contrastive objectives, or multi-turn rewards to train models to better uncover user intent \citep{wucollabllm, zhangmodeling, chen2024learning}.
\begin{figure}[h]
    \centering
    \includegraphics[width=\linewidth]{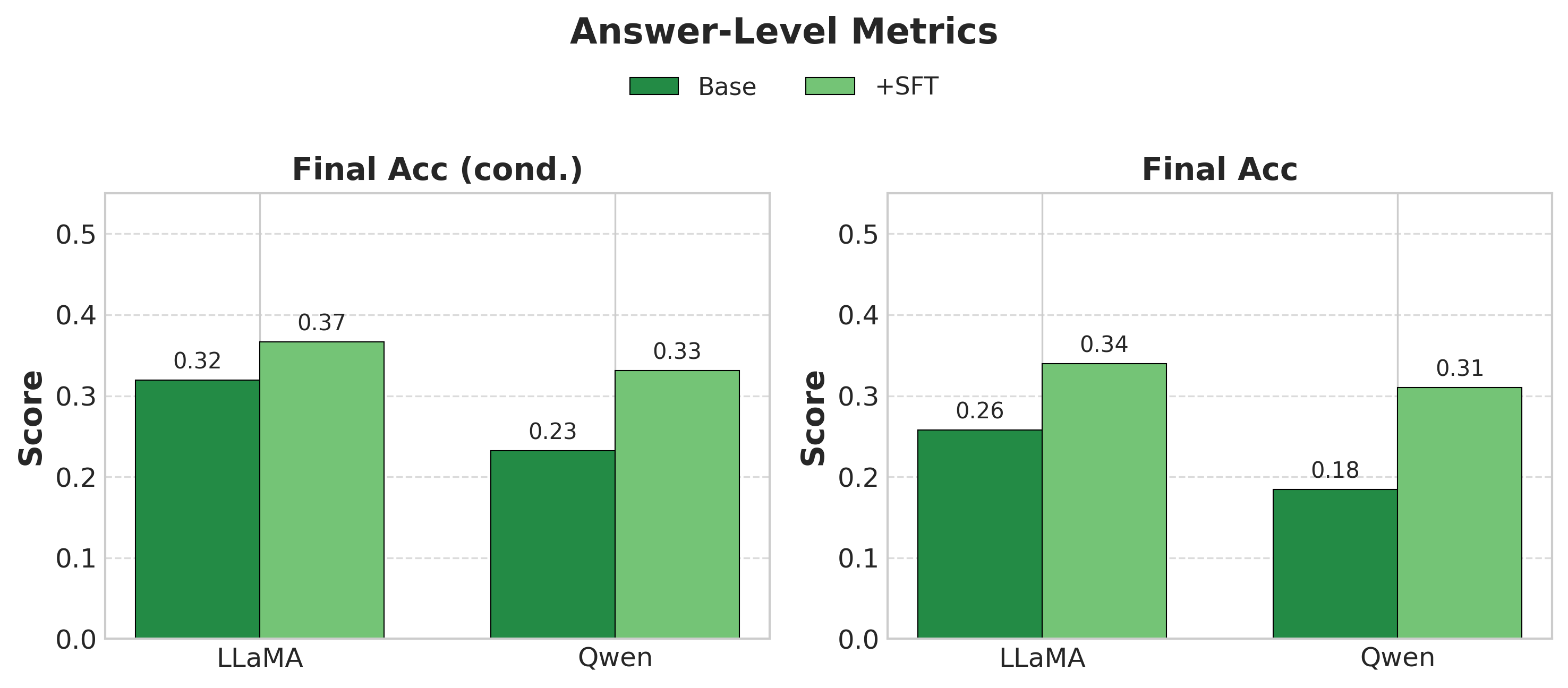}
    \caption{Answer-level metrics (Final Acc cond.\ and Final Acc) for LLaMA and Qwen before and after SFT. }
    \label{fig:answer-level-metrics}
        \vspace{-0.5cm}
\end{figure}

\subsection{Decomposing multi-turn QA performance}

Our work is also related to prior efforts that treat ambiguity resolution as a multi-stage interactive process rather than a single generation problem. For example, Zhang and Choi~\citep{zhang2025clarify} study when a system should ask for clarification, while Abg-CoQA~\citep{guo2021abg} separates ambiguity clarification from clarification-based question answering. More broadly, recent dialogue evaluation work argues that multi-turn systems should be analyzed at both the turn and dialogue levels, rather than only through end-to-end success~\citep{acikgoz2025td}.

However, these settings do not by themselves reveal which component is the main bottleneck in underspecified multi-turn QA: choosing the wrong next action, or answering incorrectly even after the correct action is taken. Our work focuses on this distinction explicitly. We decompose performance into clarification policy, which captures the ask vs. answer decision, and post-clarification answering, which captures whether the model produces the correct final answer once the missing information is available. By reporting final answer accuracy conditioned on taking the correct action, we isolate failures in post-clarification answering from failures in clarification policy.
\begin{figure}[h]
    \centering
    \includegraphics[width=\linewidth]{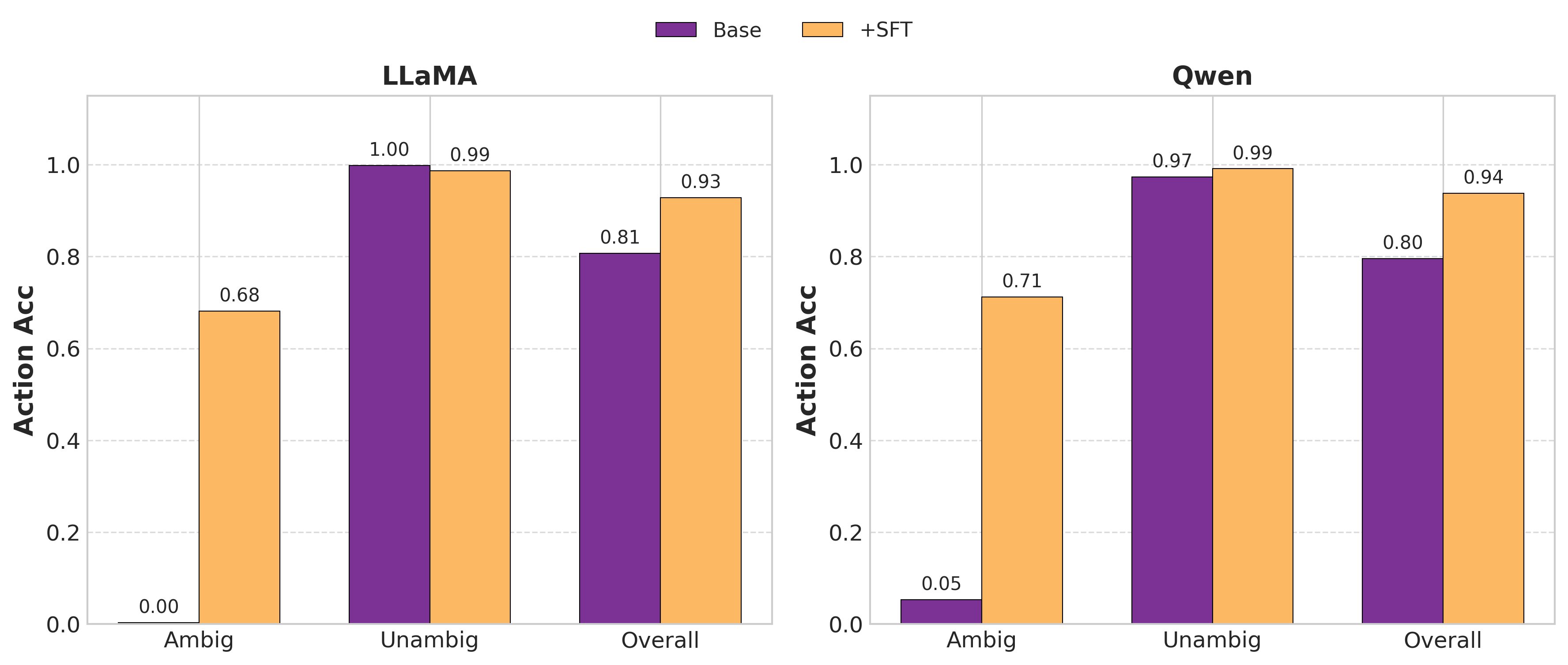}
    \caption{Action accuracy breakdown by query type (Ambig, Unambig, Overall) for LLaMA (left) and Qwen (right). Base models fail completely on ambiguous queries but achieve high accuracy after SFT.}
    \label{fig:action-acc-breakdown}
        \vspace{-0.5cm}
\end{figure}
\section{Experimental Setup}
\subsection{Dataset}
We conduct our experiments on PACIFIC \citep{deng2022pacific}, a benchmark for multi-turn conversational question answering in the financial domain. We use PACIFIC because it contains naturally occurring multi-turn interactions with both ambiguous and unambiguous user turns, allowing us to study not only whether a model chooses to ask for clarification, but also whether it answers correctly after ambiguity is resolved. This makes it well suited to our goal of separating clarification policy from post-clarification answering. It provides a realistic setting in which ambiguity arises both from individual user queries and from conversational history, such as follow-up references and underspecified context.
\begin{figure}[h]
    \centering
    \includegraphics[width=\linewidth]{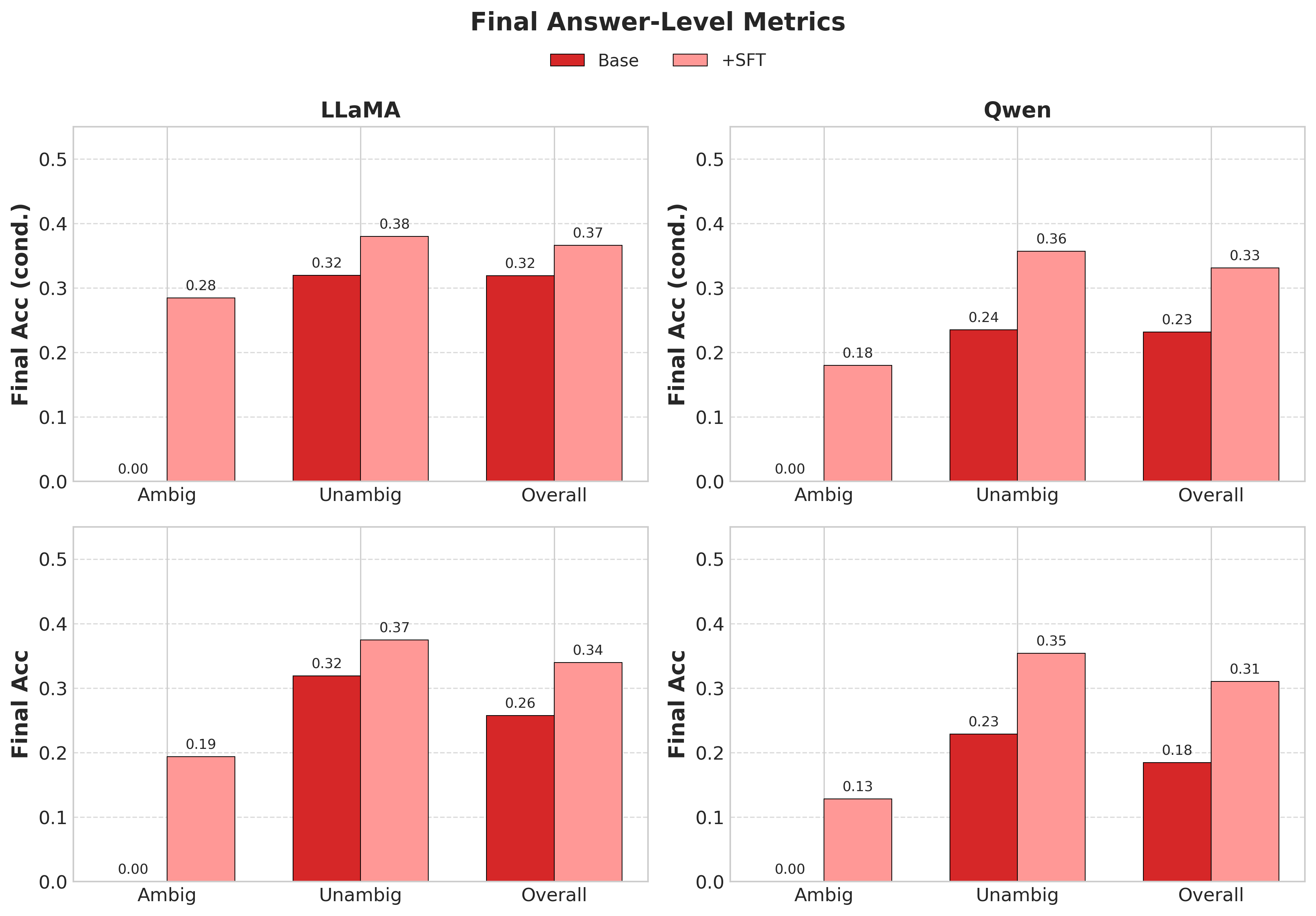}
    \caption{Final answer-level metrics breakdown by query type. }
    \label{fig:final-acc-breakdown}
        \vspace{-0.5cm}
\end{figure}
\subsection{Evaluation Metrics}
We evaluate model performance using both \textbf{action-level} and \textbf{final-answer} metrics. Action-level metrics assess whether the model chooses the correct next step action: either answering directly or asking a clarifying question. We report \emph{action accuracy}, \emph{weighted F1}, and \emph{macro F1}.
Final-answer metrics assess whether the model ultimately produces the correct answer. We report two versions. \emph{Final Acc} measures the proportion of examples for which the model both follows the gold interaction path and produces the correct final answer. \emph{Final Acc (cond.)} measures final answer accuracy conditioned on the model taking the correct action. This conditional metric is central to our analysis because it isolates failures in post-clarification answering from failures in clarification policy.

We report all metrics separately for ambiguous turns, unambiguous turns, and the full evaluation set. Action-level metrics capture whether the model responds appropriately to ambiguity, while Final Acc (cond.) reveals whether the model can answer correctly once the missing information has been supplied.
\subsection{User Simulation and Answer Evaluation}
Evaluating multi-turn clarification behavior requires both simulating user responses to model-generated clarifying questions and judging whether the model's final answer is correct. We use GPT-4o for both steps.

When the model asks a clarifying question, we prompt GPT-4o to simulate the user's reply. The simulator is given the original ambiguous question, the gold clarifying question, the gold clarification response, and the model's generated clarifying question. If the model's question is semantically aligned with the gold clarification, the simulator returns the corresponding clarification response; otherwise, it returns \texttt{irrelevant}.

To evaluate final answers, we use GPT-4o as a binary judge of semantic correctness. The judge is given the conversation context, the gold answer, and the model's final response, and outputs only \texttt{yes} or \texttt{no}. This procedure allows for semantic equivalence beyond exact string match, including paraphrases and minor formatting differences. The full prompts for user simulation and answer evaluation are provided in the Appendix.

\subsection{Models and Training}
We fine-tune two instruction-tuned models: Meta-Llama-3-8B-Instruct and Qwen2.5-7B-Instruct. All experiments use full-parameter fine-tuning with DeepSpeed ZeRO-3 for memory efficiency. We train with a learning rate of $1\times10^{-5}$, cosine learning rate decay, and a warmup ratio of 0.1. Models are trained for 8 epochs with a per-device batch size of 2 and gradient accumulation of 2, corresponding to an effective batch size of 32 on 8 GPUs. We use BF16 mixed precision and a maximum sequence length of 16{,}000 tokens. All experiments are implemented using LLaMA-Factory \citep{zheng2024llamafactory}.

\subsection{Data Augmentation}
We study several forms of synthetic augmentation using GPT-4o-generated continuations of the original training dialogues. We vary the number of augmented turns appended to each dialogue, thereby changing the synthetic trajectory length. Second, for turn augmentation, we control the proportion of clarification-seeking turns in the generated continuations through a clarification rate parameter. Third, we also explore augmentation with GPT-4o-generated reasoning traces appended to each turn.
This setup allows us to examine how the number and composition of augmented turns, as well as the use of reasoning traces, affect clarification policy and post-clarification answering. The full augmentation prompts are provided in the Appendix.
\section{Results}
\subsection{SFT Improves Clarification Policy More Than Post-Clarification Answering}
We first compare the original instruction-tuned models with their supervised fine-tuned counterparts. Table~\ref{tab:clarification-results-expanded} and Figure~\ref{fig:action-level-metrics} show that SFT yields large improvements in overall action-level performance for both LLaMA and Qwen.
In particular, SFT improves Action Accuracy and Weighted F1 by more than 10 points for both models, and also yields large gains in Macro F1 (0.42 for LLaMA and 0.40 for Qwen). These improvements are driven primarily by better handling of ambiguous turns. As shown in Figure~\ref{fig:action-acc-breakdown}, before fine-tuning, the models largely default to answering directly, which leads to high action accuracy on unambiguous turns but near-zero accuracy on ambiguous turns. After SFT, action accuracy on ambiguous turns rises to 0.68 for LLaMA and 0.71 for Qwen, while accuracy on unambiguous turns remains high, decreasing by only 0.01 and 0.02, respectively. As a result, the overall gain in action-level performance comes mainly from learning when clarification is needed, without substantially harming behavior on turns that should be answered directly.

Answer-level metrics also improve after SFT, but much less dramatically than action-level metrics. As shown in Figure~\ref{fig:answer-level-metrics}, \emph{Final Acc (cond.)} improves 0.05 for LLaMA and 0.1 for Qwen, while \emph{Final Acc} increases by 0.08 for LLaMA and 0.13 for Qwen. These improvements indicate that SFT helps not only the clarification policy but also the model's ability to produce correct final answers. However, the gains remain modest compared to the much larger improvements in action-level performance. (especially on ambiguous queries). \emph{Final Acc (cond.)} is especially informative because it measures final answer correctness conditioned on the model taking the correct next-step action. It therefore better isolates post-clarification answering from clarification-policy errors. By contrast, \emph{Final Acc} is affected by both action selection and answer correctness, and is therefore expected to benefit more directly from improvements in action accuracy. This explains why \emph{Final Acc} shows relatively larger gains than \emph{Final Acc (cond.)} after SFT.

Figure~\ref{fig:final-acc-breakdown} further breaks down these metrics by query type. Note that  \emph{Final Acc} is guaranteed to be no smaller than \emph{Final Acc (cond.) }, and the gap depends on action correctness. On unambiguous turns, \emph{Final Acc} and \emph{Final Acc (cond.) } are nearly identical, reflecting the fact that action accuracy is already very high in this category. On ambiguous turns, however, \emph{Final Acc (cond.)} remains noticeably higher than \emph{Final Acc} after SFT: 0.28 vs.\ 0.19 for LLaMA and 0.18 vs.\ 0.13 for Qwen. We also find that, even after the model takes the correct action on ambiguous turns, final answer accuracy remains relatively low, indicating that post-clarification answering remains the main challenge.

When comparing answer-level metrics across query types, we find that performance is substantially higher on unambiguous turns than on ambiguous turns. This is expected: on ambiguous turns, even if the model takes the correct action by asking for clarification, it may still fail to identify the right missing information and therefore ask the wrong clarifying question. As a result, taking the correct action is much easier than resolving the ambiguity and produce the correct final answer.

Overall, these results suggest that supervised fine-tuning rapidly teaches clarification policy, but does not equally solve the harder problem of answering correctly after clarification.
\subsection{Additional Analyses}
Appendix~\ref{sec:additional-experiments} reports several additional experiments. We find that synthetic turn augmentation provides only modest gains and quickly saturates, varying the clarification rate has limited impact, and results are broadly consistent across different augmentation models. We also find that naive reasoning trace augmentation does not solve the bottleneck and can even degrade performance. Together, these results suggest that the central challenge is not only learning to clarify, but improving post-clarification answering.

\section*{Limitations}

Our experiments are conducted exclusively on PACIFIC, a financial-domain benchmark whose questions are grounded primarily in tables and support mostly extractive or calculation-based answers. As a result, our findings may not directly generalize to other domains or to more open-ended generative tasks, where ambiguity and post-clarification reasoning may take different forms. That said, PACIFIC remains a useful testbed because it contains naturally occurring multi-turn ambiguity and allows us to cleanly study the interaction between clarification behavior and final-answer correctness.


\bibliography{example_paper}
\bibliographystyle{icml2026}

\newpage
\appendix
\onecolumn

\begin{figure*}[htbp]
\centering
\begin{tcolorbox}[
    colback=blue!5!white,
    colframe=blue!75!black,
    title={\textbf{PACIFIC Dataset Example 1}},
    fonttitle=\bfseries,
    width=\textwidth,
    enhanced,
    drop shadow
]

\begin{tcolorbox}[
    colback=gray!10!white,
    colframe=gray!50!black,
    title={\textbf{Financial Context}},
    fonttitle=\bfseries\small,
    width=\textwidth,
    enhanced
]
\textbf{Product Categories:} Drinkable Kefir, European-style soft cheeses, Cream and other, ProBugs, Other Dairy, Frozen Kefir

\vspace{8pt}
\textbf{Net Sales by Category (in thousands):}
\begin{center}
\small
\begin{tabular}{@{}lcccc@{}}
\toprule
\textbf{Category} & \textbf{2019 (\$)} & \textbf{2019 (\%)} & \textbf{2018 (\$)} & \textbf{2018 (\%)} \\
\midrule
Drinkable Kefir & 71,822 & 77\% & 78,523 & 76\% \\
Cheese & 11,459 & 12\% & 11,486 & 11\% \\
Cream and other & 4,228 & 4\% & 5,276 & 5\% \\
ProBugs Kefir & 2,780 & 3\% & 2,795 & 3\% \\
Other dairy & 1,756 & 2\% & 3,836 & 4\% \\
Frozen Kefir & 1,617 & 2\% & 1,434 & 1\% \\
\midrule
\textbf{Total} & \textbf{93,662} & \textbf{100\%} & \textbf{103,350} & \textbf{100\%} \\
\bottomrule
\end{tabular}
\end{center}
\end{tcolorbox}

\vspace{10pt}

\begin{tcolorbox}[
    colback=green!5!white,
    colframe=green!50!black,
    title={\textbf{Multi-turn Conversations}},
    fonttitle=\bfseries\small,
    width=\textwidth,
    enhanced
]
\small
\begin{tcolorbox}[colback=white, colframe=blue!30!black, boxrule=0.5pt, left=2pt, right=2pt, top=2pt, bottom=2pt]
\textbf{Turn 1:} \textcolor{blue}{Q:} "How many product categories are available?" \\
\phantom{\textbf{Turn 1:}} \textcolor{red}{A:} "6"
\end{tcolorbox}

\begin{tcolorbox}[colback=yellow!20!white, colframe=orange!50!black, boxrule=1pt, left=2pt, right=2pt, top=2pt, bottom=2pt]
\textbf{Turn 2:} \textcolor{blue}{Q:} "What is net sales from cream and other?" \\
\phantom{\textbf{Turn 2:}} \textcolor{red}{A:} "Which year are you asking about?"
\end{tcolorbox}

\begin{tcolorbox}[colback=white, colframe=blue!30!black, boxrule=0.5pt, left=2pt, right=2pt, top=2pt, bottom=2pt]
\textbf{Turn 3:} \textcolor{blue}{Q:} "in 2018 and 2019 respectively." \\
\phantom{\textbf{Turn 3:}} \textcolor{red}{A:} "5,276", "4,228"
\end{tcolorbox}

\begin{tcolorbox}[colback=white, colframe=blue!30!black, boxrule=0.5pt, left=2pt, right=2pt, top=2pt, bottom=2pt]
\textbf{Turn 4:} \textcolor{blue}{Q:} "How about ProBugs Kefir?" \\
\phantom{\textbf{Turn 4:}} \textcolor{red}{A:} "2,795", "2,780"
\end{tcolorbox}

\begin{tcolorbox}[colback=white, colframe=blue!30!black, boxrule=0.5pt, left=2pt, right=2pt, top=2pt, bottom=2pt]
\textbf{Turn 5:} \textcolor{blue}{Q:} "How about cheese?" \\
\phantom{\textbf{Turn 5:}} \textcolor{red}{A:} "11,486", "11,459"
\end{tcolorbox}

\begin{tcolorbox}[colback=white, colframe=blue!30!black, boxrule=0.5pt, left=2pt, right=2pt, top=2pt, bottom=2pt]
\textbf{Turn 6:} \textcolor{blue}{Q:} "What is the change in the net sales for that category between 2018 and 2019?" \\
\phantom{\textbf{Turn 6:}} \textcolor{red}{A:} "-27"
\end{tcolorbox}

\begin{tcolorbox}[colback=yellow!20!white, colframe=orange!50!black, boxrule=1pt, left=2pt, right=2pt, top=2pt, bottom=2pt]
\textbf{Turn 7:} \textcolor{blue}{Q:} "What is the percentage change in net sales in that period?" \\
\phantom{\textbf{Turn 7:}} \textcolor{red}{A:} "Which product category are you asking about?"
\end{tcolorbox}

\begin{tcolorbox}[colback=white, colframe=blue!30!black, boxrule=0.5pt, left=2pt, right=2pt, top=2pt, bottom=2pt]
\textbf{Turn 8:} \textcolor{blue}{Q:} "Frozen Kefir." \\
\phantom{\textbf{Turn 8:}} \textcolor{red}{A:} "12.76"
\end{tcolorbox}
\end{tcolorbox}

\vspace{10pt}

\end{tcolorbox}
\caption{Example 1 from PACIFIC dataset.}
\label{fig:dataset_example}
\end{figure*}

\begin{figure*}[htbp]
\centering
\begin{tcolorbox}[
    colback=blue!5!white,
    colframe=blue!75!black,
    title={\textbf{PACIFIC Dataset Example 2}},
    fonttitle=\bfseries,
    width=\textwidth,
    enhanced,
    drop shadow
]

\begin{tcolorbox}[
    colback=gray!10!white,
    colframe=gray!50!black,
    title={\textbf{Financial Context}},
    fonttitle=\bfseries\small,
    width=\textwidth,
    enhanced
]
\textbf{Stock-Based Compensation Summary (in thousands):}

\vspace{8pt}
\textbf{Stock-based compensation by type of award:}
\begin{center}
\small
\begin{tabular}{@{}lccc@{}}
\toprule
\textbf{Type of Award} & \textbf{2019} & \textbf{2018} & \textbf{2017} \\
\midrule
Stock options & \$648 & \$1,353 & \$2,705 \\
Stock awards & 14,882 & 10,445 & 11,421 \\
Employee stock purchase rights\textsuperscript{(1)} & 999 & 5,240 & 3,077 \\
\midrule
\textbf{Total} & \textbf{\$16,529} & \textbf{\$17,038} & \textbf{\$17,203} \\
\bottomrule
\end{tabular}
\end{center}

\vspace{8pt}
\textbf{Stock-based compensation by category of expense:}
\begin{center}
\small
\begin{tabular}{@{}lccc@{}}
\toprule
\textbf{Category} & \textbf{2019} & \textbf{2018} & \textbf{2017} \\
\midrule
Cost of revenue & \$1,500 & \$1,602 & \$1,362 \\
 Sales and marketing & 5,765 & 5,667 & 6,075 \\
Research and development & 6,039 & 6,631 & 6,343 \\
General and administrative & 3,225 & 3,138 & 3,423 \\
\midrule
\textbf{Total} & \textbf{\$16,529} & \textbf{\$17,038} & \textbf{\$17,203} \\
\bottomrule
\end{tabular}
\end{center}

\vspace{4pt}
\small
\textbf{Note:} \textsuperscript{(1)} Amount for 2018 includes \$4.1 million of accelerated stock-based compensation expense.
\end{tcolorbox}

\vspace{4pt}

\begin{tcolorbox}[
    colback=green!5!white,
    colframe=green!50!black,
    title={\textbf{Multi-turn Conversations}},
    fonttitle=\bfseries\small,
    width=\textwidth,
    enhanced
]
\small
\begin{tcolorbox}[colback=yellow!20!white, colframe=orange!50!black, boxrule=1pt, left=2pt, right=2pt, top=2pt, bottom=2pt]
\textbf{Turn 1:} \textcolor{blue}{Q:} "What is the amount of employee stock purchase rights including accelerated stock-based compensation expense?" \\
\phantom{\textbf{Turn 1:}} \textcolor{red}{A:} "Which year are you asking about?"
\end{tcolorbox}

\begin{tcolorbox}[colback=white, colframe=blue!30!black, boxrule=0.5pt, left=2pt, right=2pt, top=2pt, bottom=2pt]
\textbf{Turn 2:} \textcolor{blue}{Q:} "At the end of 2019." \\
\phantom{\textbf{Turn 2:}} \textcolor{red}{A:} "999"
\end{tcolorbox}

\begin{tcolorbox}[colback=white, colframe=blue!30!black, boxrule=0.5pt, left=2pt, right=2pt, top=2pt, bottom=2pt]
\textbf{Turn 3:} \textcolor{blue}{Q:} "How about at the end of 2018?" \\
\phantom{\textbf{Turn 3:}} \textcolor{red}{A:} "5,240"
\end{tcolorbox}

\begin{tcolorbox}[colback=white, colframe=blue!30!black, boxrule=0.5pt, left=2pt, right=2pt, top=2pt, bottom=2pt]
\textbf{Turn 4:} \textcolor{blue}{Q:} "How about 2017?" \\
\phantom{\textbf{Turn 4:}} \textcolor{red}{A:} "3,077"
\end{tcolorbox}

\begin{tcolorbox}[colback=white, colframe=blue!30!black, boxrule=0.5pt, left=2pt, right=2pt, top=2pt, bottom=2pt]
\textbf{Turn 5:} \textcolor{blue}{Q:} "What is the percentage change in total stock base compensation by type of award between 2018 and 2019?" \\
\phantom{\textbf{Turn 5:}} \textcolor{red}{A:} "-2.99"
\end{tcolorbox}

\begin{tcolorbox}[colback=white, colframe=blue!30!black, boxrule=0.5pt, left=2pt, right=2pt, top=2pt, bottom=2pt]
\textbf{Turn 6:} \textcolor{blue}{Q:} "What is its total amount between 2017 to 2019?" \\
\phantom{\textbf{Turn 6:}} \textcolor{red}{A:} "50770"
\end{tcolorbox}

\begin{tcolorbox}[colback=yellow!20!white, colframe=orange!50!black, boxrule=1pt, left=2pt, right=2pt, top=2pt, bottom=2pt]
\textbf{Turn 7:} \textcolor{blue}{Q:} "What is the percentage change in stock based compensation between 2018 and 2019?" \\
\phantom{\textbf{Turn 7:}} \textcolor{red}{A:} "What type of stock based compensation are you asking about?"
\end{tcolorbox}

\begin{tcolorbox}[colback=white, colframe=blue!30!black, boxrule=0.5pt, left=2pt, right=2pt, top=2pt, bottom=2pt]
\textbf{Turn 8:} \textcolor{blue}{Q:} "Sales and marketing." \\
\phantom{\textbf{Turn 8:}} \textcolor{red}{A:} "1.73"
\end{tcolorbox}
\end{tcolorbox}

\vspace{2pt}

\end{tcolorbox}
\caption{Example 2 from PACIFIC dataset.}
\label{fig:dataset_example2}
\end{figure*}

\begin{figure*}[htbp]
\begin{tcolorbox}[
    colback=blue!5!white,
    colframe=blue!75!black,
    title={\textbf{User Simulation Prompt}},
    fonttitle=\bfseries,
    width=\textwidth,
    enhanced,
    drop shadow
]

\textbf{Task:} You are simulating a user in a question-answering dialogue.

\vspace{8pt}

\textbf{Context:}
\begin{itemize}
\item In the last turn, you asked an ambiguous question: \texttt{[ambiguous question]}
\item The assistant should have asked you this clarifying question: \texttt{[gold clarifying question]}
\item You should have answered: \texttt{[clarifying answer]}
\item So that the assistant could give you the final answer: \texttt{[final assistant gold answer]}
\item Now, in the actual conversation, the assistant asked: \texttt{[assistant clarifying question]}
\end{itemize}

\vspace{8pt}

\textbf{Decision Rule:}
\begin{tcolorbox}[colback=yellow!10!white, colframe=orange!50!black, boxrule=1pt]
You need to determine whether the assistant asked the correct clarifying question — i.e., whether the \texttt{[assistant clarifying question]} is semantically similar to the \texttt{[gold clarifying question]}.

\textbf{If it is:} return \texttt{[clarifying answer]}

\textbf{If it is not:} return \texttt{"irrelevant"}
\end{tcolorbox}

\vspace{8pt}

\textbf{Template:}
\begin{tcolorbox}[colback=gray!10!white, colframe=gray!50!black]
\small
\texttt{Now, simulate the user response for the following conversation:}\\
\texttt{\{conversation\}}\\
\texttt{[gold clarifying question] \{gold\_clarifying\_question\}}\\
\texttt{[clarifying answer] \{clarifying\_answer\}}\\
\texttt{[final assistant gold answer] \{final\_assistant\_gold\_answer\}}\\
\texttt{[assistant clarifying question] \{assistant\_clarifying\_question\}}\\
\texttt{as a simulated user, your response:}
\end{tcolorbox}

\end{tcolorbox}
\end{figure*}

\begin{figure*}[htbp]
\begin{tcolorbox}[
    colback=green!5!white,
    colframe=green!75!black,
    title={\textbf{Clarification Turn Generation Prompt}},
    fonttitle=\bfseries,
    width=\textwidth,
    enhanced,
    drop shadow
]

\textbf{Task:} Given a conversation between a user and an assistant, simulate the next turn conversation where the user asks an ambiguous question and the assistant asks a clarifying question.

\vspace{8pt}

\textbf{Output Format:}
\begin{tcolorbox}[colback=gray!10!white, colframe=gray!50!black]
\small
\texttt{User: <ambiguous user question>}\\
\texttt{Assistant: <asking for clarifying question>}\\
\texttt{User: <User clarification>}\\
\texttt{Assistant: <assistant response after user clarification>}
\end{tcolorbox}

\vspace{8pt}

\textbf{Requirements:}
\begin{itemize}
\item The user question must be related to the information provided in the conversation and genuinely require clarification.
\item The response must contain two turns following the format above.
\end{itemize}

\vspace{8pt}

\textbf{Template:}
\begin{tcolorbox}[colback=gray!10!white, colframe=gray!50!black]
\small
\texttt{Now, generate based on this history:}\\
\texttt{\{previous\_conversations\}}
\end{tcolorbox}

\end{tcolorbox}
\end{figure*}

\begin{figure*}[htbp]
\begin{tcolorbox}[
    colback=purple!5!white,
    colframe=purple!75!black,
    title={\textbf{Direct Answering Turn Generation Prompt}},
    fonttitle=\bfseries,
    width=\textwidth,
    enhanced,
    drop shadow
]

\textbf{Task:} Given a conversation between a user and an assistant, simulate the next turn conversation where the user asks a question that is \textit{not} ambiguous and the assistant provides the answer rather than asking a clarifying question.

\vspace{8pt}

\textbf{Output Format:}
\begin{tcolorbox}[colback=gray!10!white, colframe=gray!50!black]
\small
\texttt{User: <User question>}\\
\texttt{Assistant: <Answer to the user question>}
\end{tcolorbox}

\vspace{8pt}

\textbf{Requirements:}
\begin{itemize}
\item The user question must be related to the information provided in the conversation.
\item The question must be unambiguous and not require further clarification from the user.
\item The response must contain one turn following the format above.
\end{itemize}

\vspace{8pt}

\textbf{Template:}
\begin{tcolorbox}[colback=gray!10!white, colframe=gray!50!black]
\small
\texttt{Now, generate based on this history:}\\
\texttt{\{previous\_conversations\}}
\end{tcolorbox}

\end{tcolorbox}
\end{figure*}

\begin{figure*}[htbp]
\begin{tcolorbox}[
    colback=red!5!white,
    colframe=red!75!black,
    title={\textbf{Answer Correctness Judge Prompt}},
    fonttitle=\bfseries,
    width=\textwidth,
    enhanced,
    drop shadow
]

\textbf{Task:} Evaluate whether the assistant's answer is semantically correct given the dialogue context and gold answer.

\vspace{8pt}

\textbf{Template:}
\begin{tcolorbox}[colback=gray!10!white, colframe=gray!50!black]
\small
\texttt{You are a helpful assistant evaluating whether the assistant answer in the last turn is semantically correct given the dialogue and gold answer.}\\[4pt]
\texttt{conversation: \{conversation\}}\\
\texttt{gold answer: \{gold\_answer\}}\\
\texttt{assistant answer: \{assistant\_answer\}}\\[4pt]
\texttt{Please respond with only "yes" or "no".}
\end{tcolorbox}

\end{tcolorbox}
\end{figure*}

\begin{figure*}[htbp]
\begin{tcolorbox}[
    colback=orange!5!white,
    colframe=orange!75!black,
    title={\textbf{Reasoning Trace Generation Prompt}},
    fonttitle=\bfseries,
    width=\textwidth,
    enhanced,
    drop shadow
]

\textbf{Task:} Analyze a conversation and generate reasoning traces explaining why the assistant chose to answer directly (\texttt{[answer]}) or ask a clarifying question (\texttt{[ask]}) at each turn.

\vspace{8pt}

\textbf{Output Format:} For each assistant response, append a \texttt{[reason]} tag with a concise one-sentence explanation.

\vspace{8pt}

\textbf{Example Output:}
\begin{tcolorbox}[colback=gray!10!white, colframe=gray!50!black]
\small
\texttt{User: How many product categories are available?}\\
\texttt{Assistant: [answer] 6}\\
\texttt{[reason] The context lists six distinct categories.}\\[4pt]
\texttt{User: What is net sales from cream and other?}\\
\texttt{Assistant: [ask] Which year are you asking about?}\\
\texttt{[reason] The data includes different values for both 2018 and 2019, but the user did not specify a year.}
\end{tcolorbox}

\vspace{8pt}

\textbf{Template:}
\begin{tcolorbox}[colback=gray!10!white, colframe=gray!50!black]
\small
\texttt{Now analyze this example, make sure that your reasoning is concise with only 1 sentence:}\\
\texttt{input text: \{input\_text\}}\\[4pt]
\texttt{output text:}
\end{tcolorbox}

\end{tcolorbox}
\end{figure*}

\newpage
\clearpage
\section{Additional Experiments}
\label{sec:additional-experiments}

\subsection{Synthetic Turn Augmentation Gives Modest Gains with Diminishing Returns}
We next examine whether adding synthetic conversational turns to the training data improves downstream performance. Tables~\ref{tab:llama_turns_grouped} and \ref{tab:qwen_turns_grouped} report results when we vary the number of GPT-4o-generated turns appended to each training dialogue, while fixing the clarification rate at 0.7. For both model families, turn augmentation improves performance over SFT without augmentation. For example, with one additional turn, overall Final Acc increases from 0.34 to 0.38 for LLaMA and from 0.31 to 0.34 for Qwen.

\begin{figure}[h]
    \centering
\includegraphics[width=\linewidth]{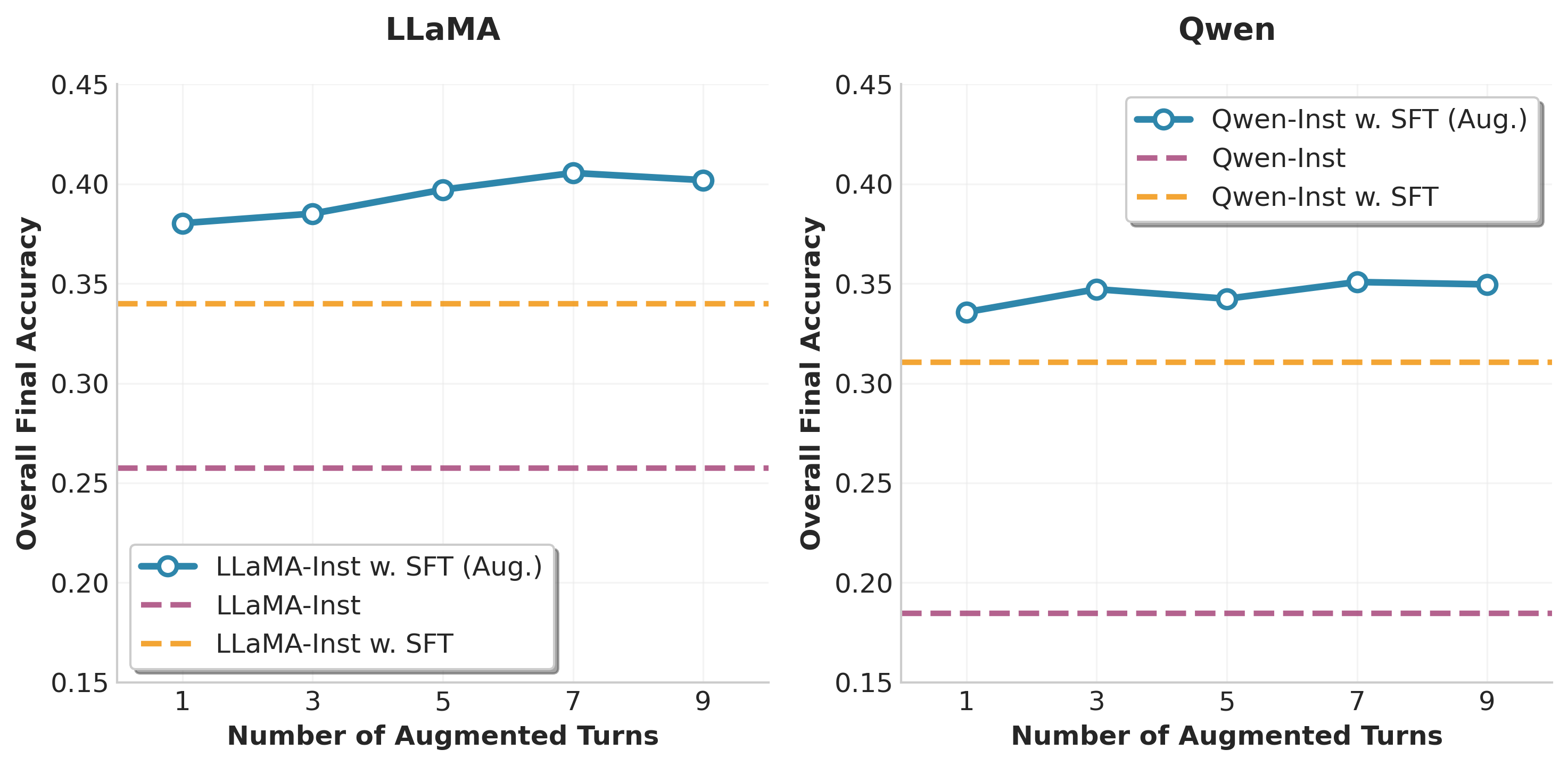}
    \caption{Effect of the number of augmented turns on overall final accuracy for Qwen (left) and LLaMA (right). Adding synthetic turns improves performance over SFT without augmentation, but gains diminish quickly and largely saturate after one additional turn. Blue lines show SFT with augmentation; orange and pink dashed lines show SFT without augmentation and instruction-tuned baselines, respectively.}
    \label{fig:plot-turns}
\end{figure}

However, these gains do not continue to scale strongly as more synthetic turns are added. Although performance improves slightly in some settings, the gains are modest and show clear diminishing returns (Figure~\ref{fig:plot-turns}). This pattern suggests that simply increasing the amount of synthetic multi-turn supervision is not sufficient to substantially close the remaining gap in post-clarification answering. In other words, longer synthetic trajectories help, but they do not fundamentally remove the main bottleneck identified above.

\subsection{Performance Is Relatively Robust to Clarification Rate}

\begin{figure}[h]
    \centering
\includegraphics[width=\linewidth]{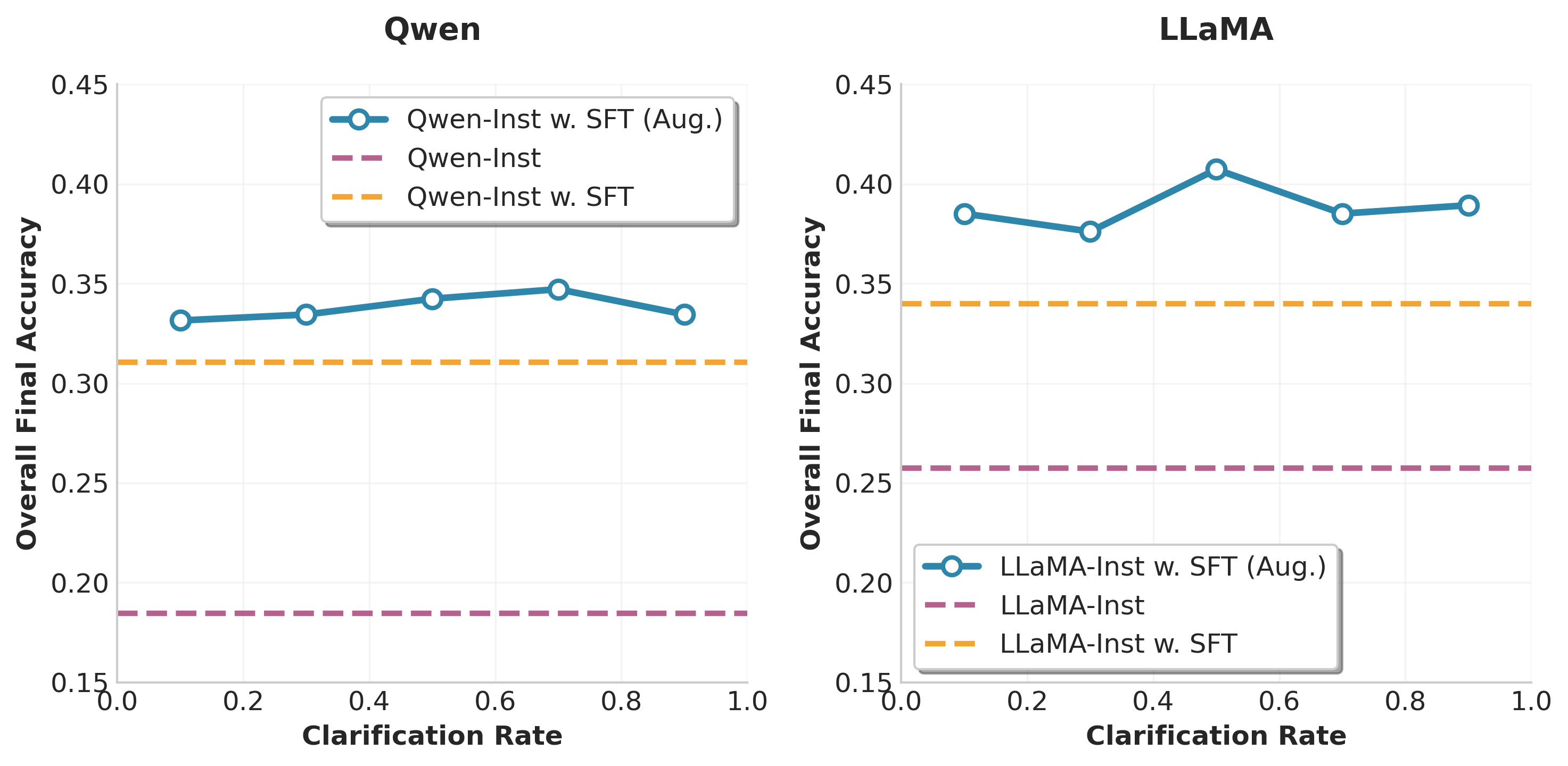}
    \caption{Effect of clarification rate in augmented data on overall final accuracy for Qwen (left) and LLaMA (right). Models are trained with 3 augmented turns at varying clarification rates (0.1--0.9). Performance is relatively robust to the proportion of clarifying turns, with only modest variations across the range. Blue lines show SFT with augmentation; orange and pink dashed lines show SFT without augmentation and instruction-tuned baselines, respectively.}
    \label{fig:plot-clarification-rate}
\end{figure}

We further study whether the composition of synthetic turn augmentation matters by varying the proportion of clarification-seeking turns while fixing the number of added turns to 3. Tables~\ref{tab:llama-clarification-rate} and \ref{tab:qwen-clarification-rate} show that overall performance varies only modestly across clarification rates for both LLaMA and Qwen (Figure~\ref{fig:plot-clarification-rate}). Although some settings perform slightly better than others, no strong monotonic trend emerges.

This result suggests that synthetic augmentation is not highly sensitive to the exact proportion of clarification turns in the training data. Increasing the share of clarification-heavy interactions alone does not consistently yield large gains.

\subsection{Downstream Performance Is Not Highly Sensitive to the Augmentation Model}

We also test whether downstream performance depends strongly on the model used to generate synthetic turn augmentations. We compare data generated by Claude-sonnet-4, Gemini-Flash, Gemini-Pro, and O3. As shown in Tables~\ref{tab:reasoning-aug-all-models} and \ref{tab:qwen-reasoning-aug}, all augmentation models yield similar downstream results, with only modest variation across generators.

\subsection{Naive Reasoning-Trace Augmentation Does Not Solve the Bottleneck}

We also investigate whether augmenting the training data with GPT-4o-generated reasoning traces improves post-clarification answering. We find that training directly on unfiltered reasoning traces degrades performance relative to training on the original dialogue data alone (Table~\ref{tab:llama-reasoning-conditioned}). In contrast, when reasoning traces are provided at inference time to the same fine-tuned model, performance improves substantially.

\begin{table*}[ht]
\centering
\small
\begin{tabular}{llccc|cc}
\toprule
\textbf{Model} & \textbf{Category} & \multicolumn{3}{c|}{\textbf{Action-Level Metrics}} & \multicolumn{2}{c}{\textbf{Final Answer-Level Metrics}} \\
\cmidrule(r){3-5} \cmidrule(l){6-7}
& & \textbf{Action Acc} & \textbf{Weighted F1} & \textbf{Macro F1} & \textbf{Final Acc (cond.)} & \textbf{Final Acc} \\
\midrule
\multirow{3}{*}{\textbf{Reasoning-Cond}}
  & Unambig  & 0.9694 & 0.9845 & 0.4922 & 0.5657 & 0.5484 \\
  & Ambig    & 0.9563 & 0.9776 & 0.4888 & 0.1275 & 0.1219 \\
  & Overall  & 0.9669 & 0.9674 & 0.9484 & 0.4823 & 0.4663 \\
\midrule
\multirow{3}{*}{\textbf{Reasoning-Uncond}}
  & Unambig  & 0.8830 & 0.9379 & 0.4689 & 0.3249 & 0.2869 \\
  & Ambig    & 0.6906 & 0.8170 & 0.4085 & 0.1403 & 0.0969 \\
  & Overall  & 0.8460 & 0.8507 & 0.7679 & 0.2959 & 0.2503 \\
\bottomrule
\end{tabular}
\caption{Comparison of LLaMA models trained with reasoning traces, evaluated with (Reasoning-Cond) and without (Reasoning-Uncond) conditioning on reasoning during inference. Training on unfiltered reasoning traces degrades performance, but conditioning on reasoning at inference time provides a substantial boost.}
\label{tab:llama-reasoning-conditioned}
\end{table*}
\begin{table*}[ht]
\centering
\small
\begin{tabular}{llccc|cc}
\toprule
\textbf{Model} & \textbf{Category} & \multicolumn{3}{c|}{\textbf{Action-Level Metrics}} & \multicolumn{2}{c}{\textbf{Final Answer-Level Metrics}} \\
\cmidrule(r){3-5} \cmidrule(l){6-7}
& & \textbf{Action Acc} & \textbf{Weighted F1} & \textbf{Macro F1} & \textbf{Final Acc (cond.)} & \textbf{Final Acc} \\
\midrule
\multirow{3}{*}{\textbf{LLaMA}}
  & Unambig  & 0.9985 & - & - & 0.3194 & 0.3189 \\
  & Ambig    & 0.0031 & - &-  & 0.0000 & 0.0000 \\
  & Overall  & 0.8069 & 0.7223 & 0.4496 & 0.3192 & 0.2575 \\
\midrule
\multirow{3}{*}{\textbf{LLaMA+SFT}}
  & Unambig  & 0.9866 & - & - & 0.3799 & 0.3748 \\
  & Ambig    & 0.6813 & - & - & 0.2844 & 0.1938 \\
  & Overall  & 0.9278 & 0.9234 & 0.8704 & 0.3664 & 0.3400 \\
\midrule
\multirow{3}{*}{\textbf{Qwen}}
  & Unambig  & 0.9732 & - & - & 0.2351 & 0.2288 \\
  & Ambig    & 0.0531 & - & - & 0.0000 & 0.0000 \\
  & Overall  & 0.7960 & 0.7323 & 0.4881 & 0.2320 & 0.1847 \\
\midrule
\multirow{3}{*}{\textbf{Qwen+SFT}}
  & Unambig  & 0.9911 & - & - & 0.3571 & 0.3539 \\
  & Ambig    & 0.7125 & - & - & 0.1798 & 0.1281 \\
  & Overall  & 0.9374 & 0.9339 & 0.8883 & 0.3312 & 0.3105 \\
\bottomrule
\end{tabular}
\caption{Comparison of LLaMA and Qwen models (with and without SFT) on ambiguous and unambiguous questions. }
\label{tab:clarification-results-expanded}
\end{table*}

\begin{table*}[ht]
\centering
\small
\begin{tabular}{llccc|cc}
\toprule
\textbf{Model} & \textbf{Category} & \textbf{Action Acc} & \textbf{Weighted F1} & \textbf{Macro F1} & \textbf{Final Acc (cond.)} & \textbf{Final Acc} \\
\midrule
\multirow{3}{*}{\textbf{LLaMA}}
  & Unambig  & 0.9978 & 0.9989 & 0.4994 & 0.4190 & 0.4180 \\
  & Ambig    & 0.5156 & 0.6804 & 0.3402 & 0.1091 & 0.0563 \\
  & Overall  & 0.9049 & 0.8927 & 0.8103 & 0.3850 & 0.3484 \\
\midrule
\multirow{3}{*}{\textbf{Qwen}}
  & Unambig  & 0.9911 & 0.9955 & 0.4978 & 0.3564 & 0.3532 \\
  & Ambig    & 0.7844 & 0.8792 & 0.4396 & 0.1355 & 0.1063 \\
  & Overall  & 0.9513 & 0.9494 & 0.9158 & 0.3213 & 0.3057 \\
\bottomrule
\end{tabular}
\caption{Performance of LLaMA and Qwen models when directly conditioned on GPT-4o-generated reasoning traces. The relatively low final accuracy indicates that a large portion of the generated reasoning traces are noisy or unhelpful.}
\label{tab:model-ambig-breakdown}
\end{table*}

\begin{table*}[ht]
\centering
\small
\begin{tabular}{llccc|cc}
\toprule
\textbf{Model} & \textbf{Category} & \textbf{Action Acc} & \textbf{Weighted F1} & \textbf{Macro F1} & \textbf{Final Acc (cond.)} & \textbf{Final Acc} \\
\midrule
\multirow{3}{*}{\textbf{Claude}}
  & Unambig  & 0.9814 & - & - & 0.4131 & 0.4054 \\
  & Ambig    & 0.7969 & - & - & 0.2941 & 0.2344 \\
  & Overall  & 0.9458 & 0.9444 & 0.9085 & 0.3938 & 0.3724 \\
\midrule
\multirow{3}{*}{\textbf{Gemini Flash}}
  & Unambig  & 0.9806 & - & - & 0.4020 & 0.3942 \\
  & Ambig    & 0.8063 & - & - & 0.3062 & 0.2469 \\
  & Overall  & 0.9471 & 0.9458 & 0.9110 & 0.3863 & 0.3658 \\
\midrule
\multirow{3}{*}{\textbf{Gemini Pro}}
  & Unambig  & 0.9881 & - & - & 0.3967 & 0.3920 \\
  & Ambig    & 0.7000 & - & - & 0.3080 & 0.2156 \\
  & Overall  & 0.9326 & 0.9288 & 0.8797 & 0.3839 & 0.3580 \\
\midrule
\multirow{3}{*}{\textbf{O3}}
  & Unambig  & 0.9836 & - & - & 0.4121 & 0.4054 \\
  & Ambig    & 0.7250 & - & - & 0.2888 & 0.2094 \\
  & Overall  & 0.9338 & 0.9308 & 0.8842 & 0.3937 & 0.3676 \\
\bottomrule
\end{tabular}
\caption{Performance LLaMA fine-tuned on turn augmented generated by Claude, Gemini Flash, Gemini Pro, and O3. }
\label{tab:reasoning-aug-all-models}
\end{table*}
\begin{table*}[ht]
\centering
\small
\begin{tabular}{llccc|cc}
\toprule
\textbf{Model} & \textbf{Category} & \textbf{Action Acc} & \textbf{Weighted F1} & \textbf{Macro F1} & \textbf{Final Acc (cond.)} & \textbf{Final Acc} \\
\midrule
\multirow{3}{*}{\textbf{Claude}}
  & Unambig  & 0.9881 & - & - & 0.3808 & 0.3763 \\
  & Ambig    & 0.8219 & - & - & 0.2281 & 0.1875 \\
  & Overall  & 0.9561 & 0.9549 & 0.9257 & 0.3556 & 0.3400 \\
\midrule
\multirow{3}{*}{\textbf{Gemini Flash}}
  & Unambig  & 0.9724 & - & - & 0.3785 & 0.3681 \\
  & Ambig    & 0.8156 & - & - & 0.1762 & 0.1438 \\
  & Overall  & 0.9422 & 0.9414 & 0.9046 & 0.3448 & 0.3249 \\
\midrule
\multirow{3}{*}{\textbf{Gemini Pro}}
  & Unambig  & 0.9918 & - & - & 0.3719 & 0.3689 \\
  & Ambig    & 0.7594 & - & - & 0.2181 & 0.1656 \\
  & Overall  & 0.9471 & 0.9446 & 0.9073 & 0.3482 & 0.3297 \\
\midrule
\multirow{3}{*}{\textbf{O3}}
  & Unambig  & 0.9829 & -& - & 0.3889 & 0.3823 \\
  & Ambig    & 0.8000 & - & -& 0.1797 & 0.1438 \\
  & Overall  & 0.9477 & 0.9463 & 0.9114 & 0.3549 & 0.3363 \\
\bottomrule
\end{tabular}
\caption{Performance Qwen fine-tuned on turn augmented generated by Claude, Gemini Flash, Gemini Pro, and O3.}
\label{tab:qwen-reasoning-aug}
\end{table*}

\begin{table*}[ht]
\centering
\small
\begin{tabular}{llccc|cc}
\toprule
\textbf{Turns} & \textbf{Category} & \textbf{Action Acc} & \textbf{Weighted F1} & \textbf{Macro F1} & \textbf{Final Acc (cond.)} & \textbf{Final Acc} \\
\midrule
\multirow{3}{*}{\textbf{1}}
  & Unambig  & 0.9873 & - & - & 0.4189 & 0.4136 \\
  & Ambig    & 0.7938 &- & - & 0.3031 & 0.2406 \\
  & Overall  & 0.9501 & 0.9484 & 0.9146 & 0.4003 & 0.3803 \\
\midrule
\multirow{3}{*}{\textbf{3}}
  & Unambig  & 0.9776 & - & - & 0.4306 & 0.4210 \\
  & Ambig    & 0.7813 & - & - & 0.3000 & 0.2344 \\
  & Overall  & 0.9398 & 0.9383 & 0.8983 & 0.4097 & 0.3851 \\
\midrule
\multirow{3}{*}{\textbf{5}}
  & Unambig  & 0.9784 & - & - & 0.4455 & 0.4359 \\
  & Ambig    & 0.7875 & - & - & 0.2976 & 0.2344 \\
  & Overall  & 0.9416 & 0.9402 & 0.9015 & 0.4217 & 0.3971 \\
\midrule
\multirow{3}{*}{\textbf{7}}
  & Unambig  & 0.9881 & - & - & 0.4480 & 0.4426 \\
  & Ambig    & 0.7469 & - & - & 0.3347 & 0.2500 \\
  & Overall  & 0.9416 & 0.9390 & 0.8980 & 0.4307 & 0.4055 \\
\midrule
\multirow{3}{*}{\textbf{9}}
  & Unambig  & 0.9806 & - & - & 0.4415 & 0.4329 \\
  & Ambig    & 0.7875 &- & - & 0.3452 & 0.2719 \\
  & Overall  & 0.9434 & 0.9419 & 0.9042 & 0.4260 & 0.4019 \\
\bottomrule
\end{tabular}
\caption{Performance of LLaMA model with increasing augmented turns.}
\label{tab:llama_turns_grouped}
\end{table*}

\begin{table*}[ht]
\centering
\small
\begin{tabular}{llccc|cc}
\toprule
\textbf{Turns} & \textbf{Category} & \textbf{Action Acc} & \textbf{Weighted F1} & \textbf{Macro F1} & \textbf{Final Acc (cond.)} & \textbf{Final Acc} \\
\midrule
\multirow{3}{*}{\textbf{1}}
  & Unambig  & 0.9769 & - & - & 0.3875 & 0.3785 \\
  & Ambig    & 0.7625 & - & - & 0.2049 & 0.1563 \\
  & Overall  & 0.9356 & 0.9337 & 0.8905 & 0.3588 & 0.3357 \\
\midrule
\multirow{3}{*}{\textbf{3}}
  & Unambig  & 0.9821 & - & - & 0.3991 & 0.3920 \\
  & Ambig    & 0.7188 & - & - & 0.2217 & 0.1594 \\
  & Overall  & 0.9314 & 0.9283 & 0.8800 & 0.3727 & 0.3472 \\
\midrule
\multirow{3}{*}{\textbf{5}}
  & Unambig  & 0.9799 & - & - & 0.3833 & 0.3756 \\
  & Ambig    & 0.7625 & - & - & 0.2664 & 0.2031 \\
  & Overall  & 0.9380 & 0.9360 & 0.8940 & 0.3650 & 0.3424 \\
\midrule
\multirow{3}{*}{\textbf{7}}
  & Unambig  & 0.9799 & - & - & 0.3939 & 0.3860 \\
  & Ambig    & 0.7969 & - & - & 0.2549 & 0.2031 \\
  & Overall  & 0.9446 & 0.9433 & 0.9067 & 0.3713 & 0.3508 \\
\midrule
\multirow{3}{*}{\textbf{9}}
  & Unambig  & 0.9836 & - & - & 0.3977 & 0.3912 \\
  & Ambig    & 0.7594 & - & - & 0.2305 & 0.1750 \\
  & Overall  & 0.9404 & 0.9382 & 0.8973 & 0.3717 & 0.3496 \\
\bottomrule
\end{tabular}
\caption{Performance of Qwen model with increasing augmented turns.}
\label{tab:qwen_turns_grouped}
\end{table*}

\begin{table*}[ht]
\centering
\small
\begin{tabular}{llccc|cc}
\toprule
\textbf{Portion} & \textbf{Category} & \textbf{Action Acc} & \textbf{Weighted F1} & \textbf{Macro F1} & \textbf{Final Acc (cond.)} & \textbf{Final Acc} \\
\midrule
\multirow{3}{*}{\textbf{0.1}}
  & Unambig  & 0.9873 & - & - & 0.4302 & 0.4247 \\
  & Ambig    & 0.7531 & - & - & 0.2905 & 0.2188 \\
  & Overall  & 0.9422 & 0.9398 & 0.8995 & 0.4087 & 0.3851 \\
\midrule
\multirow{3}{*}{\textbf{0.3}}
  & Unambig  & 0.9769 & - & - & 0.4157 & 0.4061 \\
  & Ambig    & 0.7969 & - & - & 0.3137 & 0.2500 \\
  & Overall  & 0.9422 & 0.9410 & 0.9031 & 0.3991 & 0.3761 \\
\midrule
\multirow{3}{*}{\textbf{0.5}}
  & Unambig  & 0.9791 & - & - & 0.4543 & 0.4449 \\
  & Ambig    & 0.7719 & - & - & 0.3239 & 0.2500 \\
  & Overall  & 0.9392 & 0.9374 & 0.8966 & 0.4337 & 0.4073 \\
\midrule
\multirow{3}{*}{\textbf{0.7}}
  & Unambig  & 0.9776 & - & - & 0.4306 & 0.4210 \\
  & Ambig    & 0.7813 & - & - & 0.3000 & 0.2344 \\
  & Overall  & 0.9398 & 0.9383 & 0.8983 & 0.4097 & 0.3851 \\
\midrule
\multirow{3}{*}{\textbf{0.9}}
  & Unambig  & 0.9821 &- & - & 0.4378 & 0.4300 \\
  & Ambig    & 0.7563 & - & - & 0.2893 & 0.2188 \\
  & Overall  & 0.9386 & 0.9364 & 0.8943 & 0.4147 & 0.3893 \\
\bottomrule
\end{tabular}
\caption{Performance of LLaMA models trained with 3 augmented turns and varying clarification rate.}
\label{tab:llama-clarification-rate}
\end{table*}

\begin{table*}[ht]
\centering
\small
\begin{tabular}{llccc|cc}
\toprule
\textbf{Portion} & \textbf{Category} & \textbf{Action Acc} & \textbf{Weighted F1} & \textbf{Macro F1} & \textbf{Final Acc (cond.)} & \textbf{Final Acc} \\
\midrule
\multirow{3}{*}{\textbf{0.1}}
  & Unambig  & 0.9806 &- & - & 0.3792 & 0.3718 \\
  & Ambig    & 0.7500 & - & - & 0.2167 & 0.1625 \\
  & Overall  & 0.9362 & 0.9339 & 0.8902 & 0.3541 & 0.3315 \\
\midrule
\multirow{3}{*}{\textbf{0.3}}
  & Unambig  & 0.9918 & - & - & 0.3809 & 0.3778 \\
  & Ambig    & 0.7000 & - & - & 0.2188 & 0.1531 \\
  & Overall  & 0.9356 & 0.9317 & 0.8843 & 0.3576 & 0.3345 \\
\midrule
\multirow{3}{*}{\textbf{0.5}}
  & Unambig  & 0.9769 &- & - & 0.3913 & 0.3823 \\
  & Ambig    & 0.7563 & - & - & 0.2314 & 0.1750 \\
  & Overall  & 0.9344 & 0.9324 & 0.8881 & 0.3664 & 0.3424 \\
\midrule
\multirow{3}{*}{\textbf{0.7}}
  & Unambig  & 0.9821 & - & - & 0.3991 & 0.3920 \\
  & Ambig    & 0.7188 & - & - & 0.2217 & 0.1594 \\
  & Overall  & 0.9314 & 0.9283 & 0.8800 & 0.3727 & 0.3472 \\
\midrule
\multirow{3}{*}{\textbf{0.9}}
  & Unambig  & 0.9821 & - &- & 0.3771 & 0.3703 \\
  & Ambig    & 0.7938 & - & - & 0.2323 & 0.1844 \\
  & Overall  & 0.9458 & 0.9444 & 0.9082 & 0.3537 & 0.3345 \\
\bottomrule
\end{tabular}
\caption{Performance of Qwen models trained with 3 augmented turns and varying clarification rate.}
\label{tab:qwen-clarification-rate}
\end{table*}

\end{document}